# Accurate Discharge Coefficient Prediction of Streamlined Weirs by Coupling Linear Regression and Deep Convolutional Gated Recurrent Unit

Weibin Chen[a], Danial Sharifrazi[b], Guoxi Liang[c], Shahab S. Band[d], Kwok Wing Chau[e], Amir Mosavi[f,g,*]

[a] *College of Computer Science and Artificial Intelligence, Wenzhou University, Wenzhou 325035, Zhejiang, China*
[b] *Department of Computer Engineering, Shiraz Branch, Islamic Azad University, Shiraz, Iran*
[c] *Department of Artificial Intelligence, Wenzhou Polytechnic, Wenzhou 325035, China*
[d] *Future Technology Research Center, College of Future, National Yunlin University of Science and Technology, Yunlin County, Taiwan*
[e] *Department of Civil and Environmental Engineering, Hong Kong Polytechnic University, Hong Kong, China*
[f] *John von Neumann Faculty of Informatics, Obuda University, Budapest, Hungary*
[g] *Institute of Information Engineering, Automation and Mathematics, Slovak University of Technology in Bratislava, Slovakia*

**Corresponding author:** Amir Mosavi (amir.mosavi@uni-obuda.hu)

**Abstract:** Streamlined weirs, as a nature-inspired type of weirs, has gained tremendous attention among hydraulic engineers mainly due to their well-known performance with high discharge coefficient. Computational fluid dynamic (CDF) is considered as a robust tool to predict discharge coefficient. To bypass the computational cost of CFD-based assessment, the present study proposes data-driven modelling techniques, as an alternative to CFD simulation, to predict discharge coefficient based on an experimental dataset. To this end, after splitting the dataset by k-fold cross-validation technique, the performance assessment of classical and hybrid machine-deep learning (ML-DL) algorithms is undertaken. Amongst ML techniques, linear regression (LR), random forest (RF), support vector machine (SVM), k-nearest neighbours (KNN), and decision tree (DT) algorithms are studied . In the context of DL, long short-term memory (LSTM), convolutional neural network (CNN), gated recurrent unit (GRU) and their hybrid forms such as LSTM-GRU, CNN-LSTM and CNN-GRU techniques are compared by different error metrics. It is found that the proposed three-layer hierarchical DL algorithm consisting of a convolutional layer coupled with two subsequent GRU levels, which is also hybridized by LR method (i.e., LR-CGRU), leads to lower error metrics. This paper paves the way for data-driven modelling of streamlined weirs.

## 1. Introduction

Weirs are the most useful and common hydraulic structures, which are applied in various usages such as irrigation networks, sewage networks and water supply systems (Abdollahi et al., 2017). According to the crest type, main weir groups are classified into sharp-, broad-, and short-crested weirs. Circular-crested, overflow (ogee) and streamlined weirs are special kinds of short-crested weirs (Bagheri & Kabiri-Samani, 2020a). Streamlined weirs, as a nature-inspired type of weirs, has gained tremendous attention among hydraulic engineers due to their well-known performance with high discharge coefficient, overflow stability behaviour and minimized fluctuation in water free surface. The general shape of streamlined weirs, which is designed according to aerofoils, is originally derived from birds' wing topology. The importance of streamlined weirs, purported to be the most state-of-the-art form of weirs, is well-documented in hydraulic engineering field (Rao & Rao, 1973; Bagheri & Kabiri-Samani, 2020). However, due to the complexity of the geometry of streamlined weir in design, this kind of weir has been paid less attention among practitioners. The estimation of the discharge coefficient of weirs is an important subject since many experimental and/or numerical researches have been undertaken recently in different types of weirs (Arvanaghi et al., 2014; Arvanaghi & Oskuei, 2013; Borghei et al., 1999; Johnson, 2000; Mahtabi & Arvanaghi, 2018; Qu et al., 2009; Rady, 2011; Tullis, 2011). For the last two decades, computational fluid dynamics (CFD) has drawn tremendous attention from both academia and industry to model problems that involve fluid domains and their corresponding boundary condition and interactions. OpenFOAM software, as an open-source toolbox, is widely used in high-fidelity computational models due to its incorporation of a vast variety of solvers compatible with different range of fluid flows. Although CFD-based performance assessment of fluid-flow phenomena leads to reliable results, it suffers from computationally demanding procedures and a requirement of profound academic knowledge in the field of fluid mechanics (Bagheri & Kabiri-Samani, 2020b, 2020a). Data-driven modelling offers a framework to assess a model as a black-box. Hence, it is possible to analyse a broader range of models and systems irrespective of the nature of the problem. In particular, ML-DL modelling is an active field of research in other engineering fields such as structural and earthquake engineering (Abasi et al., 2021; M S Barkhordari & Es-haghi, 2021; Mohammad Sadegh Barkhordari & Tehranizadeh, 2021; Esteghamati & Flint, 2021; Hariri-Ardebili & Salazar, 2020; Pourkamali-Anaraki et al., 2020; Soraghi & Huang, 2021), biomedical engineering (Alizadehsani et al., 2021; Ayoobi et al., 2021), etc. Other applications of ML-DL techniques can also be found (Aswin et al., 2018; Athira et al., 2018; Selvin et al., 2017; Vinayakumar et al., 2017).

Recently, different ML and surrogate modeling algorithms have been applied in various hydraulic engineering problems such as dams, sedimentation, spillway, etc. (Amini et al., 2021; Bhattacharya et al., 2007; Hariri-Ardebili et al., 2021; Roushangar et al., 2014; Torres-Rua et al., 2012). It is recognized that an empirical relationship for discharge coefficient based on experimental or hydraulic models faces some limitations regarding hydraulic and geometric parameters (Ebtehaj et al., 2018). The main motivation of the present study is to bypass the computational cost of discharge coefficient prediction via CFD framework by investigating the potential capability of hybrid ML-DL algorithms as an alternative to CFD-based simulations. The comparison between CFD-based discharge coefficient and the proposed data-driven techniques is also graphically illustrated in Figure 1. which is inspired from (Bagheri & Kabiri-Samani, 2020a and

2020b). The data-driven modelling part of Figure 1 will be discussed comprehensively in Sections 4 and 5.

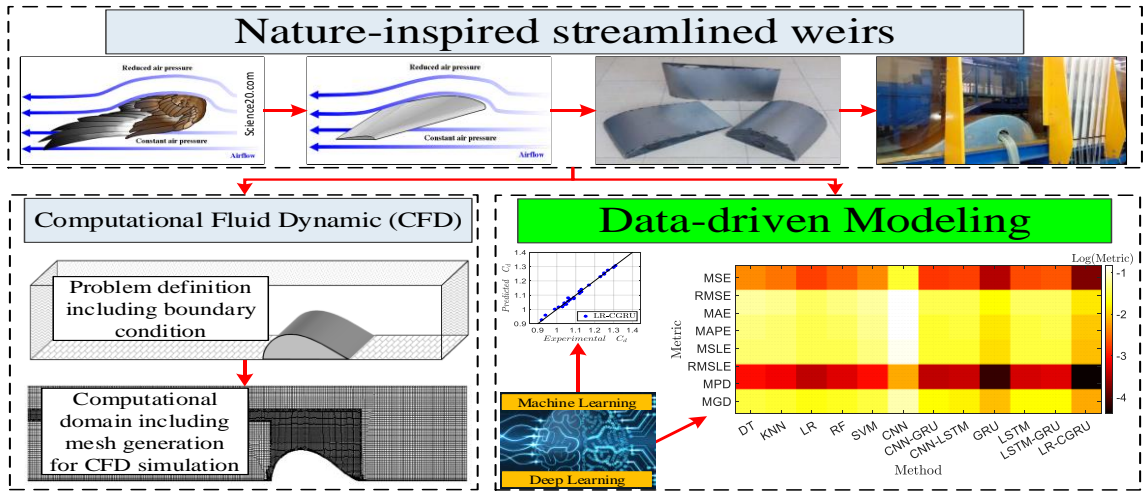

Figure 1. Data-driven discharge coefficient estimation of streamlined weirs as an alternative to CFD-based procedure

The incorporation of various geometric and hydraulic parameters affecting hydraulic operations of weirs require the application of an accurate model to determine their discharge coefficients. In this context, the need for proposing an accurate technique for the estimation of discharge coefficient is a challenging task.

In this work, a group of 12 classical and hybrid ML-DL algorithms are employed to predict the discharge coefficient of streamlined weirs based on an experimental dataset. In the following, Section 2 describes literature related to different usages of ML-DL techniques in weirs. Section 3 explains the data employed in this study. Section 4 describes the ML-DL algorithms including the proposed one. Section 5 illustrates results obtained by different data-driven techniques. Finally, in Section 6, the conclusion is presented, and future works are outlined.

## 2. Related works

The determination of discharge coefficient of weirs is the most momentous factor for the design of these hydraulic structures. Several studies were performed by using various ML-DL algorithms to predict the discharge coefficient. In this section, some of the state-of-the-art ML-DL techniques related to the estimation of the discharge coefficient are presented in Table 1 considering different weir configurations. One may note that none of the existing studies investigated the potential capability of ML-DL techniques for streamlined weirs which reflects the main motivation of the present study.

Table 1- Previous works on discharge coefficient estimation of weirs via different soft computing techniques

| Weir configuration | Soft computing technique | Reference |
|---|---|---|
| Sharp-crested weir | FFNN[1], RBNN[2] | (Bilhan et al., 2010) |
| Triangular labyrinth side weirs | ANN[3] | (Emiroglu et al., 2011) |
| Broad-crested weir | GP[4], ANN | (Salmasi et al., 2013) |
| Triangular labyrinth side weirs | MLP[5], RBNN | (Zaji & Bonakdari, 2014) |
| Side weirs | MLP | (Parsaie & Haghiabi, 2015) |
| Trapezoidal and rectangular side weirs | SVM[6] and GA[7] (SVM-GA), GEP[8] | (Roushangar et al., 2016) |
| Two-cycle labyrinth weirs | ANFIS[9], MNLR[10] | (Aydin & Kayisli, 2016) |
| Side weirs | SVM | (Azamathulla et al., 2016) |
| Triangular labyrinth weirs | SVR[11], SVR-FA[12], RSM[13], PCA[14] | (Karami et al., 2017) |
| Triangular labyrinth weirs | MLP-NN, RBNN, SVM | (Parsaie & Haghiabi, 2017) |
| Rectangular side weirs | ANFIS | (Ebtehaj et al., 2018) |
| Labyrinth weirs | ANFIS, MLP-NN | (Haghiabi et al., 2018) |
| Piano key weir | MLP, MLP-FA, MLP-PSO[15], MLP-GA, MLP-MFO[16], ANFIS, ANFIS-FA, ANFIS-PSO, ANFIS-GA, ANFIS-MFO | (Zounemat-Kermani et al., 2019) |
| Trapezoidal labyrinth weirs | MLP-NN, RBNN, SVM | (Norouzi et al., 2019) |
| Labyrinth weirs | ANFIS, ANFIS-FFA[17] | (Shafiei et al., 2020) |
| Skew side weir | MLR[18], GEP | (Mohammed & Sharifi, 2020) |
| Sharp-crested weirs | ANN, SVM, ELM[19] | (Li et al., 2021) |
| Triangular labyrinth weirs | ANFIS, ANFIS-PSO, ANFIS-FA, SVR, SVR-FA, MLP, MLP-FA, RBNN | (Mahmoud et al., 2021) |

[1] Feed forward neural network, [2] Radial basis neural networks, [3] Artificial neural network, [4] Genetic programming, [5] Multi-layer perceptron neural network, [6] Support vector machine, [7] Genetic algorithm, [8] Gene expression programming, [9] Adaptive neuro-fuzzy inference system, [10] Multiple nonlinear regression, [11] Support vector regression, [12] Firefly algorithm, [13] Response surface methodology, [14] Principal component analysis, [15] Particle swarm optimization, [16] Moth-flame optimization, [17] Neuro-fuzzy-firefly, [18] Multiple linear regression, [19] Extreme learning machine

## 3. Data description

The flow rate $Q$ over a short-crest weir are computed based on continuity and Bernoulli's equations as expressed in Equation (1):

$$Q = \frac{2}{3} C_d B \sqrt{\frac{2}{3} g} H_1^{3/2} \qquad (1)$$

where $C_d$ is weir discharge coefficient; $B$ represents weir width; $H_1 = h_1 + h_v$ describes total head; $h_1$ is upstream head over the crest; $h_v$ indicates upstream velocity head and equals to $v^2/2g$; $v$ refers to approach velocity; and $g$ denotes the acceleration due to gravity.

In this research, an experimental dataset for 120 models of streamlined weirs, which are designed based on the principle of the Joukowsky transform function, is used (Bagheri & Kabiri-Samani, 2020a). The model is graphically illustrated in Figure 2 and the related hydraulic parameters are shown in Table 2. The data consist of two groups, namely with and without base-block under streamlined weirs. In models without base-block, parameter $\beta$ is considered equal to zero. Table 2 shows 9 parameters, which are considered as model inputs in the proposed method. Besides, the discharge coefficient is the model output.

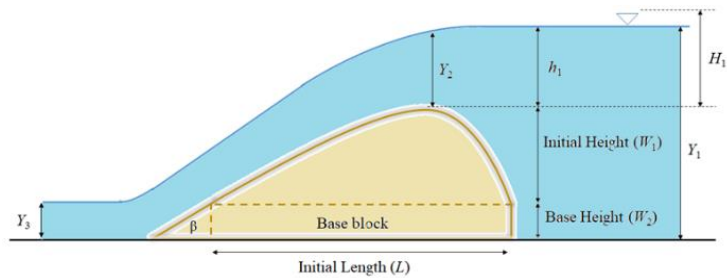

Figure 2. Schematic view of streamline weir (adapted and modified from Bagheri & Kabiri-Samani (2020a))

Table 2. Input parameters for estimating discharge coefficient

| Input parameters | Description of input parameters |
|---|---|
| $\lambda$ | relative eccentricity |
| $\beta$ | angle between the downstream slope of weirs fixed and horizontal axis |
| $L$ | initial length of the streamlined weir |
| $W$ | total weir height |
| $Q$ | flow discharge |
| $Y_1$ | upstream water depth |
| Y$_2$ | water depth at the weir crest |
| Y$_3$ | downstream flow depth |
| *h$_1$* | upstream flow depth on the weir crest |

## 4. Methods

In this section, the studied ML-DL methods are introduced in Section 4.1. Details of the implemented methods and parameters are also stated. Besides, the proposed method is introduced in detail in the following. All data-driven techniques are implemented by Python programming language. In this research, "sklearn" and "keras" packages by "tensorflow" backend are used for program development. A GPU GFORCE GTX950 with 16GB RAM DDR4 is used as the implementation hardware.

### *4.1. Machine-Deep Learning Algorithms*

With the development of ML-DL methods, a good variety of ML-DL-based models were introduced and received extended attention (see Table 1). In the present study, five classical ML techniques are applied to estimate the discharge coefficient. The performance assessment of support vector machine (SVM), random forest (RF), linear regression (LR), K-nearest neighbors (KNN) and decision tree (DT) algorithms is undertaken via error metrics. Among these ML techniques, the candidate with the highest accuracy is considered as the accepted ML technique in the present study. All model parameters of classical ML techniques are summarized in Table 3. Since the applied ML techniques are well-documented in the literature, the readers are referred to Sammut & Webb (2011) for a detailed discussion on the mentioned classical ML techniques.

Table 3. Parameter values of ML algorithms

| SVM | RF | KNN | DT |
|---|---|---|---|
| Kernel=RBF | n_estimators=100 | n_neighbors=5 | Criterion=MSE |
| Degree=3 | Criterion=MSE | Weights=Uniform | Splitter=Best |
| Gamma=Scale | min_samples_split=2 | Algorithm=Auto | min_samples_split=2 |
| Coef0=0.0 | min_samples_leaf=1 | Leaf Size=30 | min_samples_leaf=1 |
| Shrinking =True | min_weight_fraction_leaf=0. | p=2 | min_weight_fraction_leaf =0. |
| Cache size=200 | Max Features=Auto | Metric=Minkowski | |
| Epsilon=0.1 | Bootstrap=True | | |
| Tol=1e-3 | | | |
| C=1.0 | | | |

As mentioned in Section 1, the main objective of this study is to propose an accurate data-driven technique to estimate the discharge coefficient. Accordingly, we assess the capability of six classical and hybrid DL techniques in comparison to a three-layer hierarchical DL technique for the possible adaptive implementation with a successful ML technique in a state-of-the-art hydraulic engineering application. Deep neural networks (DNNs) are created from artificial neural networks (ANN). ANN usually contain few layers (shallow) whereas DNNs contain more hidden (deep) layers. With more layers, DNNs are capable of learning big data (Wang et al., 2019). Deep learning (DL) is a method that predicts results through several layers, with each layer containing the weights of a neural network (Zhao et al., 2019). As a result, it can be said that deep learning is a special kind of neural network that involves more layers. Within this framework, increasing the number of layers in DL has led to better outcomes than simple ANNs. In the context of DL, long short-term memory (LSTM) (Hochreiter & Schmidhuber, 1997), convolutional neural network (CNN) (LeCun et al., 1995), gated recurrent unit (GRU) (Cho et al., 2014) and their hybrid forms such as LSTM-GRU, CNN-LSTM and CNN-GRU techniques are analysed by different error metrics. In the following, DL techniques are introduced briefly while a detailed discussion on the proposed algorithm is provided. As a variant of recurrent neural network (RNN), LSTM has a long-term memory function that is suitable for processing important events with long intervals and delays in time series. Therefore, the neural network structure, which is primarily composed of LSTM units with memory functions, can make decisions based on previous states to adapt to various running scenarios (Guo et al., 2021). LSTM has been widely used in issues related to sequential data such as natural language processing (NLP), voice recognition, and time series analysis (Sezer & Ozbayoglu, 2018).

CNN's original idea was initially modeled on mammalian vision. This type of network is able to achieve results similar to humans in some cases and even stronger than human vision in some other cases. CNN is made up of a number of convolutional layers. From the combination of these layers of convolution, a deep neural network is formed. CNN has been widely used and achieved brilliant results in image processing, image classification and computer vision (Sammut & Webb, 2011).

Similar to LSTM, GRU is another variant of RNN. In general, two main layers are implemented in GRU. It first determines how the previous information should be passed along to the future. Next, it determines how much of the past information must be discarded in the second layer (Ayoobi et al., 2021). GRU leads to better performance for smaller and less frequent datasets in comparison to LSTM (Gruber & Jockisch, 2020). Model parameters of these classical DL techniques are summarized in Table 4.

Table 4. Parameter values of classical DL algorithms

| LSTM | CNN | GRU |
|---|---|---|
| Layers: 3 LSTM layers | Layers: 3 convolutional layers | Layers: 3 GRU layers |
| Number of neurons:50 | Number of filters: 64 | Number of neurons:50 |
| Number of epochs: 200 | Number of epochs: 200 | Number of epochs: 200 |
| Activation for all layers (except the last): ReLU | Activation for all layers (except the last): ReLU | Activation for all layers (except the last): ReLU |
| Loss function: MSE | Loss function: MSE | Loss function: MSE |
| Optimizer: Adam | Optimizer: Adam | Optimizer: Adam |
| beta1 of optimizer: 0.9 | beta1 of optimizer: 0.9 | beta1 of optimizer: 0.9 |
| beta2 of optimizer: 0.999 | beta2 of optimizer: 0.999 | beta2 of optimizer: 0.999 |
| Learning rate: 0.001 | Learning rate: 0.001 | Learning rate: 0.001 |
| | Size of kernels: 3*3 | |

Hybrid DL techniques are constructed by coupling classical DL algorithms. In this context, LSTM-GRU is developed by 2 LSTM layers and 1 GRU layer, in which the number of neurons of LSTM and GRU layers is assumed 50. The other remaining parameters are identical to LSTM and GRU parameters. In CNN-LSTM approach, one convolutional layer and 2 LSTM layers are applied whilst other remaining parameters are obtained from the classical DL. The same implementation is assumed for CNN-GRU where 1 convolutional layer and 2 GRU layers are mixed. All remaining parameters of LR-CGRU are assumed equal to those of LR, CNN and GRU algorithms.

### 4.2. Proposed Method (LR-CGRU)

The dataset is split into the "training" and "testing" groups to generate meta-inputs for the proposed algorithm. A successful out-of-sampling technique for this purpose is the k-fold cross-validation (CV) technique. In this context, by transforming the whole dataset into k mutually exclusive and collectively exhaustive subsets, only one set is used for testing and the remaining (k-1) subgroup are incorporated in the training procedure. In addition, the initial weigh assignment of ML-DL algorithms is commonly performed by a random configuration. Hence, k-fold CV technique can lead to unbiased assessment.

In the proposed ML-DL algorithm in the present study, k=5 is used for the CV tool. According to Razavi-Far et al. (2019), the predictive models are trained in "one-step-ahead" configuration.

A three-layer hierarchical DL algorithm consisting of a convolutional layer coupled with two GRU levels is introduced as the final DL algorithm, which is also hybridized by LR method as the ML technique due to its lower CV errors (detailed explanation of error metrics and their obtained values for ML-DL algorithms will be discussed in Section 5). Accordingly, LR-CGRU is the combination of LR, CNN and GRU and uses a convolutional layer as the first layer and two GRU layers subsequently in the DL phase. A graphical representation of the proposed algorithm is demonstrated in Figure 3.

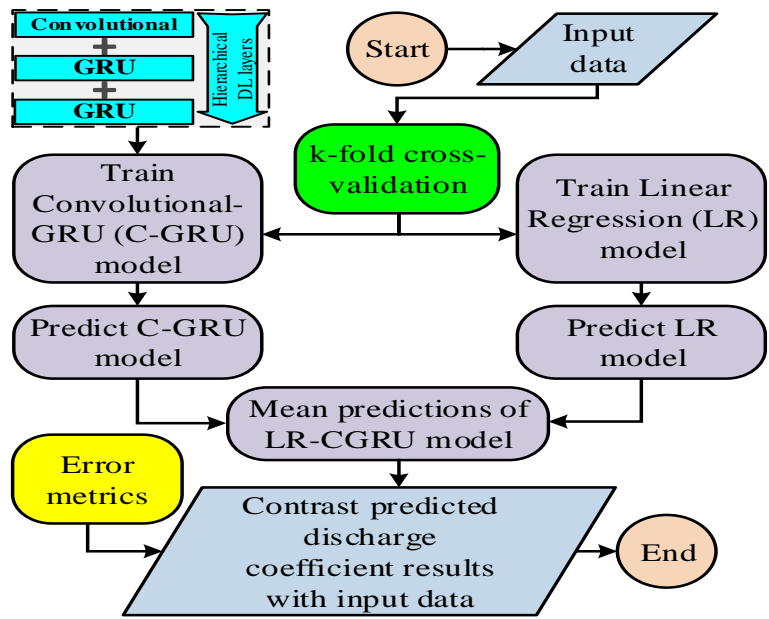

Figure 3. Flowchart of the proposed method: LR-CGRU consisting of an ML algorithm (i.e., RL) coupled with a three-layer hierarchical DL technique (i.e., CGRU)

The proposed model is trained 5 times due to the usage of 5-fold CV technique. In 5-fold CV technique, the model is trained with 80 percent of all dataset and tested by the remaining 20 percent. Accordingly, we have 5 predicted datasets by both ML and DL algorithms, in which the computed data are averaged for both ML and DL methods.

## 5-Results and discussion

### *5.1. Verification of the proposed algorithm*

In this section, at the first stage, the predicted results of all ML methods including SVM, RF, LR, KNN and DT are compared with the experimental results, which are graphically demonstrated in Figure 4(a)-(e). Intuitively, it can be observed that LR and RF methods provide better results compared to other ML techniques in terms of YY plot.

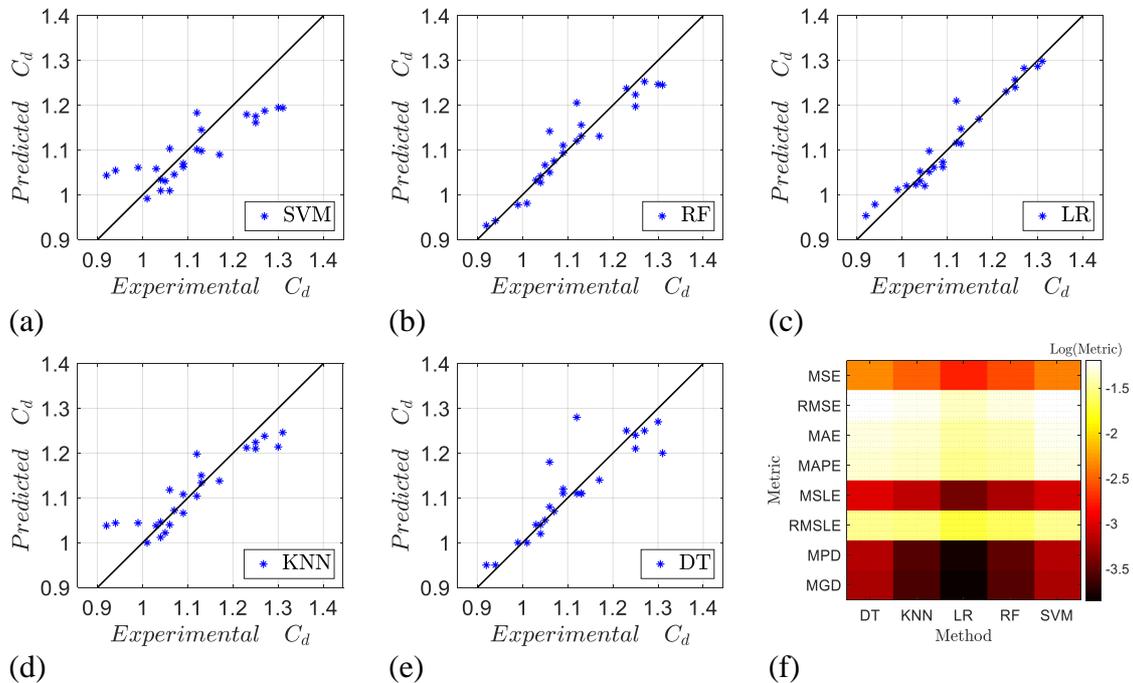

Figure 4. Comparison between the experimental data and ML methods; Figure 4(a)-(e) compare predicted discharge coefficient with experimental dataset; Figure 4(f) depicts results of 6 error metrics for all ML algorithms

An ML-DL model can be evaluated in a tricky manner. The dataset usually is split into training and testing sets. Then, the model performance is evaluated based on an error metric to specify the precision of the model. However, this technique is not reliable enough as the computed accuracy for one test set may be very different from another one. To cope with this problem, k-fold cross-validation (CV) is performed. As mentioned in

Section 4.2, 5-fold CV technique is considered for all applied ML-DL algorithms. In detail, in the first iteration, the first fold is employed to test ML-DL model and the rest of the data is considered as the training set. In the next iteration, the second fold is used as the testing set and the rest of data is employed as a training set. This procedure continues until 5 folds.

To assess the performance of each ML-DL method, eight error metrics, namely mean squared error (MSE), root mean squared error (RMSE), mean absolute error (MAE), mean absolute percentage error (MAPE), mean squared logarithmic error (MSLE), root mean squared logarithmic error (RMSLE), mean Poisson deviance (MPD), and mean Gamma deviance (MGD) are employed. These error metrics are introduced in Equations (2)-(9), respectively:

$$MSE = \frac{1}{n}\sum_{i=1}^{n}(y_i - \hat{y}_i)^2 \quad (2)$$

$$RMSE = \sqrt{\frac{1}{n}\sum_{i=1}^{n}(y_i - \hat{y}_i)^2} \quad (3)$$

$$MAE = \frac{\sum_{i=1}^{n}|y_i - \hat{y}_i|}{n} \quad (4)$$

$$MAPE = \frac{100}{n}\sum_{i=1}^{n}\left|\frac{y_i - \hat{y}_i}{y_i}\right| \quad (5)$$

$$MSLE = \frac{1}{n}\sum_{i=1}^{n}(log(y_i) - log(\hat{y}_i))^2 \quad (6)$$

$$RMSLE = \sqrt{\frac{1}{n}\sum_{i=1}^{n}(log(y_i) - log(\hat{y}_i))^2} \quad (7)$$

$$MPD = \frac{1}{n}\sum_{i=0}^{n-1}2(y_i log\left(\frac{y_i}{\hat{y}_i}\right) + \hat{y}_i - y_i) \quad (8)$$

$$MGD = \frac{1}{n}\sum_{i=0}^{n-1}2(log\left(\frac{\hat{y}_i}{y_i}\right) + \frac{y_i}{\hat{y}_i} - 1) \quad (9)$$

where $y_i$ describes real (i.e., experimental) dataset and $\hat{y}_i$ refers to predicted outputs. Figure 4(f) shows logarithmic values of the applied performance metrics for each ML method. According to Figure 4(f), linear regression, which has the darkest colour among other methods, is considered as the most successful ML technique in the present study.

In the next stage, classical DL methods (namely LSTM, GRU and CNN) and their variants (namely CNN-LSTM, GRU, LSTM, LSTM-GRU) are applied to predict discharge coefficient of streamlined weirs. The predicted outputs by the mentioned DL algorithms versus the experimental dataset are demonstrated via YY plot in Figure 5.

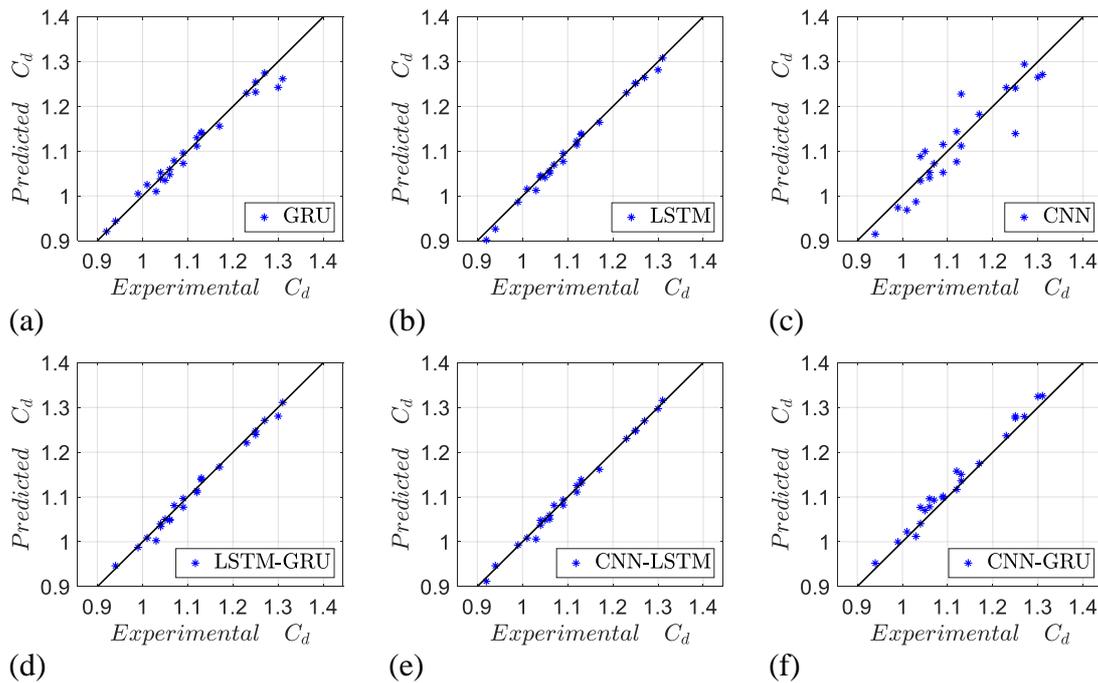

Figure 5. Comparison between the experimental dataset and derived outputs by the applied classical and hybrid DL methods

As it can be seen in the second row of Figure 5, all hybrid DL algorithms outperform the classical ones. However, to provide a robust conclusion, the mentioned eight error metrics in Figure 4(f) are applied again on which logarithmic values of error metrics are depicted in Figure 6(a). To demonstrate the potential capability of the proposed methods, the error metrics of LR-CGRU algorithm is also plotted in the last column of Figure 6(a). In general, it can be concluded that all hybrid algorithms considering both ML and DL ones which are plotted in Figures. 4(f) and 6(a), respectively, provide low error metrics. LR-CGRU not only leads to lower error considering all eight metrics, but also provides considerably lower metrics in MSE, MSLE, MPD and MGD. Moreover, YY plot for the proposed method is introduced in Figure 6(b), which highlights the superiority of LR-CGRU method. The computational cost regarding the training time of all ML-DL algorithms is also introduced in Appendix 1. As it is expected, there is a sharp distinction between computational costs of ML and DL algorithms. However, LR-CGRU provides an acceptable computational complexity compared to other classical and hybrid DL algorithms.

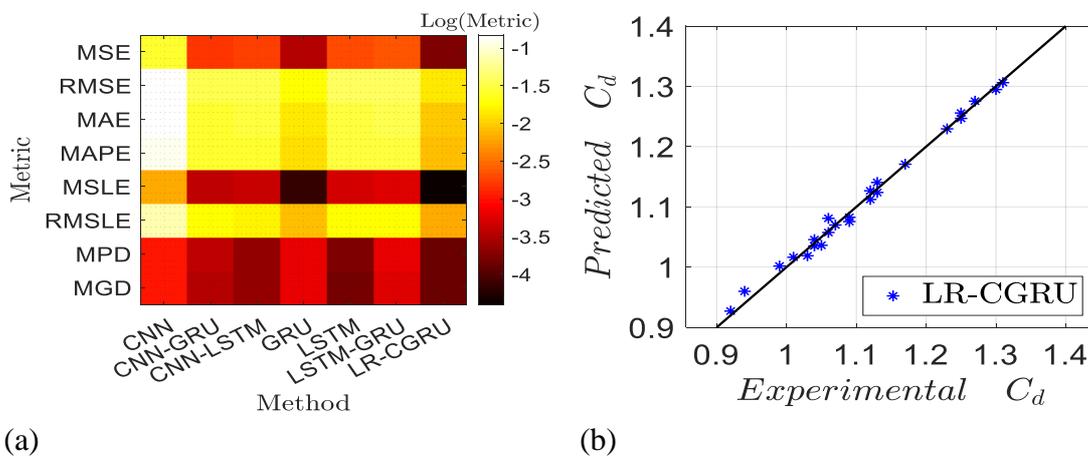

(a) (b)
Figure 6. LR-CGRU: (a); Error metrics for all DL algorithms in conjunction with LR-CGRU method (b) YY-plot for the proposed method

### 5.2. Comparison with previous works

Finally, the data-driven outputs are compared with those of previous related works. Bagheri and Kabiri-Samani (2020a) proposed an algebraic equation to compute the streamlined discharge coefficient ($C_d$) using dimensional analysis and curve-fitting tool in MATLAB as follows:

$$C_d = 1.4\lambda^{0.05}\left[\frac{h_1}{L}\frac{h_1}{W}\right]^{0.1} \quad (10)$$

Carollo and Ferro (2021) proposed a relationship between discharge $Q$ and upstream water level $h_1$, based on experimental results of Bagheri and Kabiri-Samani (2020a), as shown in Eq. (11):

$$A = a\left(\frac{h_1}{W}\right) = \frac{Q^{2/3}}{g^{1/3}b^{2/3}W} \tag{11}$$

Based on Equations (10) and (11), the coefficient $a$ was:

$$a = \frac{2}{3}C_d^{2/3} \tag{12}$$

In Carollo & Ferro (2021), according to dimensional analysis and self-similarity theory, the stage-discharge relationship was obtained as:

$$A = 0.8546\left(\frac{h_1}{W}\right)^{1.1243}\left(\frac{L}{W}\right)^{-0.1012}\left(\frac{W_1}{W}\right)^{.0412} \tag{13}$$

by combining Equations (11) and (12):

$$A = \frac{2}{3}C_d^{2/3}\frac{h_1}{W} \tag{14}$$

By substituting Equation (13) into Equation (14):

$$\frac{2}{3}C_d^{2/3}\frac{h_1}{W} = 0.8546\left(\frac{h_1}{W}\right)^{1.1243}\left(\frac{L}{W}\right)^{-0.1012}\left(\frac{W_1}{W}\right)^{.0412} \tag{15}$$

In the last step, the discharge coefficient was obtained as:

$$C_d = \left[\left[\frac{3}{2}\frac{W}{h_1}\right]\left[0.8546\left(\frac{h_1}{W}\right)^{1.1243}\left(\frac{L}{W}\right)^{-0.1012}\left(\frac{W_1}{W}\right)^{.0412}\right]\right]^{3/2} \tag{16}$$

In Figure 7, results from equations proposed by Bagheri & Kabiri-Samani (i.e., Eq. (10)) and Carollo & Ferro (i.e., Eq. (16)) are compared with those by the proposed LR-CGRU algorithm. As it can be seen, the proposed data-driven technique provides more accurate outputs than the algebraic expressions introduced by Bagheri & Kabiri-Samani (2020a) and Carollo & Ferro (2021), which highlights the superiority of ML-DL driven techniques for the prediction of discharge coefficient.

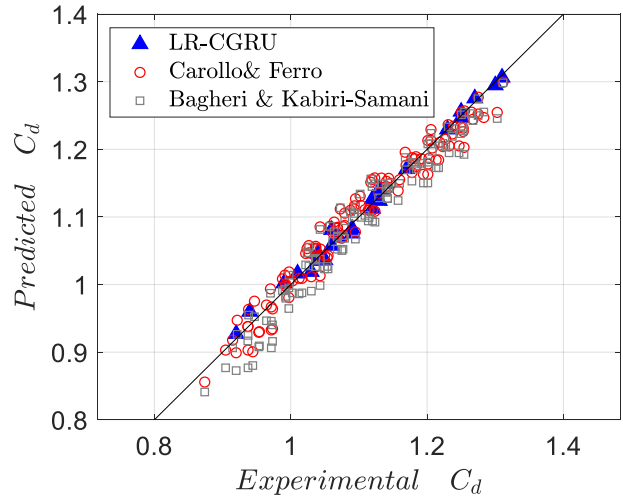

Figure 7. Comparison of LR-CGRU outputs (blue triangles) with previous works via YY-plot

## 6- Conclusion and future works

This paper aims to predict the discharge coefficient of streamlined weirs, which are known as a state-of-the-art type of weirs. As an alternative to the computational fluid dynamic procedure to predict discharge coefficient of this nature-inspired type of weirs, the potential superiority of machine learning-deep learning algorithms is investigated. Five classical machine learning techniques, namely linear regression, random forest, support vector machine, k-nearest neighbours, and decision tree, are applied. In addition, amongst deep learning algorithms, long short-term memory (LSTM), convolutional neural network (CNN) and gated recurrent unit (GRU) and their hybrid forms (i.e., LSTM-GRU, CNN-LSTM and CNN-GRU) are compared by eight different error metrics.

To enhance accuracy, a three-layer hierarchical deep learning algorithm consisting of a convolutional layer coupled with two subsequent GRU levels, which is also hybridized by the linear regression method (i.e., LR-CGRU), is proposed. In general, hybrid deep data-driven algorithms provide more accurate results than the classical ones.

Furthermore, it is clearly demonstrated that LR-CGRU technique outperforms other eleven machine-deep learning algorithms.

Finally, the superiority of the proposed data-driven technique is demonstrated by a comparative analysis between previously introduced algebraic expressions to predict discharge coefficient. Results indicate that LR-CGRU algorithm can act as an alternative tool to forecast the discharge coefficient of streamlined weirs accurately, which paves the way for data-driven modelling of streamlined weirs. Although the capabilities of twelve machine-deep learning algorithms are investigated to predict discharge coefficient, there is still a need for future studies to enhance both accuracy and efficiency of the estimation. Furthermore, investigation on the application of the proposed ML-DL algorithm in probabilistic risk assessment (Ali Amini et al., 2021; Kia et al., 2021) of streamlined weirs can be performed in future works.

**Appendix 1**:

Table 5. Computational cost of training time for all 12 ML-DL algorithms

| LR | RF | SVM | KNN | DT | |
|---|---|---|---|---|---|
| 0:00:00.003218 | 0:00:00.119285 | 0:00:00.000996 | 0:00:00.000630 | 0:00:00.000409 | |
| LSTM | CNN | LSTM-GRU | CNN-LSTM | CNN-GRU | LR-CGRU |
| 0:00:46.185944 | 0:00:04.668465 | 0:00:46.714708 | 0:00:29.236926 | 0:00:29.725064 | 0:00:29.728282 |

comprehensive review of methods, applications and data sources." Information Fusion 63 (2020): 256-272.